\documentclass[letterpaper, 10 pt, conference]{ieeeconf}  

\IEEEoverridecommandlockouts                              

\usepackage{mf-style}
\usepackage{agent-symbols}
\usepackage[utf8]{inputenc}
\usepackage{polski}

\usepackage{caption}

\usepackage{textcomp} 

\DeclareCaptionType{equ}[Eq.][]

\usepackage{amssymb}

\usepackage[]{subcaption}

\title{	
		\begin{flushleft}\vspace{-2.0cm}
			{\normalsize 2019 International Conference on Robotics and Automation (ICRA)\\\vspace{-0.3cm}
				Palais des congres de Montreal, Montreal, Canada, May 20-24, 2019}
		\end{flushleft}	
	\vspace{1.0cm}
	\LARGE \bf
	Methodology of Designing Multi-agent Robot Control Systems Utilising  Hierarchical Petri Nets
}

\author{Maksym Figat$^{1}$ and Cezary Zieli{\'{n}}ski$^{1}$
	\thanks{*This work was supported by the National Science Centre, Poland (grant number 2017/25/N/ST7/00900).}
	\thanks{$^{1}$ Warsaw University of Technology, Institute of Control and Computation Engineering,
		Nowowiejska 15/19, 00-665 Warsaw, Poland,
		{\tt\small M.Figat@ia.pw.edu.pl}}%
}

\begin{document}

	\maketitle


	\begin{abstract}
		
		A robot system is designed as a set of embodied agents.
		An embodied agent is decomposed into cooperating subsystems.
		In our previous work activities of subsystems were defined by hierarchical finite state machines. With their states activities were associated.
		In that approach communication between subsystems was treated as an implementation issue.
		This paper represents activities of a robot system using hierarchical Petri nets with conditions. Such net is created by specifying consecutive layers: multi-agent robot system layer, agent layer, subsystem layer, behaviour layer and communication layer.
		This decomposition not only organizes in a systematic manner the development of a robot system,	but also introduces a comprehensive description of concurrently acting subsystems.
		Based on those theoretical considerations, a tool was created for producing hierarchical Petri nets defining the model of a robotic system and enabling automatic generation of the robot controller code, resulting in a significant acceleration of the implementation phase.
		The capabilities of the tool are presented by the development of a robot controller performing a rudimentary task.
		
	\end{abstract}

	\section{INTRODUCTION}
	
	Specification of a robotic system requires the determination of its architecture, i.e.\ structure and style~\cite{RSAaP:08}.
	Structure pertains to the decomposition of the robotic system into subsystems and presents interconnections between them.
	Style determines the computational and communication concepts utilised to design the system.
	In many cases the definition of the structure and style of the developed systems is not obvious~\cite{RSAaP:08}, as
	often they were created without a clear architectural pattern.
	
	The control system together with the devices it controls can be composed of one or more communicating agents~\cite{Brooks:1991_intelligence,Jennings:1998_roadmap}.
	An agent is a separate system that has an internal imperative that rationally affects its surroundings based on the information collected from the environment. If the environment is physical in nature, then such an agent is called an embodied agent.
	An embodied agent affects the environment with its effectors and obtains data from it utilising its receptors~\cite{Zie:10BulPAN,Brooks:1991_intelligence,Kornuta:13_irs,ZielinskiMMAR2014}.
	An embodied agent is decomposed into a set of communicating subsystems.
	The activity of each subsystem in our previous work was defined using hierarchical finite state machines (HFSM)~\cite{Zielinski-2017-JINT,Figat:2017_RoMoCo,Zielinski:Automation2018}.
	Each state of the upper layer HFSM has a subFSM associated with it. Such subFSM specifies a single behaviour of the subsystem.
	Those HFSMs facilitate the specification of computational concepts associated with individual subsystems operating in parallel, however treated their communication as an implementation issue.
	
	Alternatively architectural style can be described by a Petri net (PN) with conditions~\cite{peterson1981petri,giraultValk2003petri,Huber91}. In that case system compliance with the requirements can be verified~\cite{Montano:2000,Billington:2011}, as Petri net verification tools exist, and above all, their ability to generate code  automatically can be utilised.  This paper presents a hierarchical Petri net \pnHierarchical\ (HPN) modelling the activities of a robotic system decomposed into five layers: 1) multi-agent robot system layer composed of agents, 2) agent layer composed of subsystems, 3) subsystem layer assigning bahaviours to subsystem ,
	4) behaviour layer representing behaviours in terms of elementary activities and 5) communication layer defining the communication model utilised within a behaviour. Out of this HPN model, robot controller code is automatically generated and verified by simulation.
	
	The structure of the paper is as follows. HPNs are presented in Sec.~\ref{sec:hierarchical-petri-net-intro}. Sec.~\ref{sec:embodied-agent} briefly reveals the concept of an embodied agent. It presents a robotic system architecture in terms of its structure and activities.
	Sec.~\ref{sec:experiment} introduces an example of a robotic system specified utilising the presented modelling method. It also discloses the method  of automatic generation of robot controller code out of a HPN.
	Sec.~\ref{sec:related-works} presents the related work and the conclusions are drawn in Sec.~\ref{sec:conclusions}.
	
	\section{HPN WITH CONDITIONS}
	\label{sec:hierarchical-petri-net-intro}
	A Petri net is a bipartite graph containing transitions \pnTransition{} and places \pnPlace{} alternatively connected by directed arcs \cite{peterson1981petri}. In a HPN  some places can be substituted by pages \pnPage{}.
	A page is a HPN with a distinguished single input place \pnInputPlace{} and single output place \pnOutputPlace{}.
	In this paper we use a HPN with conditions. In such nets with each transition a condition \pnCondition{} is associated.
	When tokens are assigned to places the net becomes a marked HPN with conditions \pnHierarchical.
	With each place a single operation $\pnOperation$ is associated.
	Places from different nets can be fused with each other.
	Fused places \pnFusionPlace{} are in principle the same place appearing in two or more nets~\cite{Huber91}.
	These places are indistinguishable from each other and thus contain the same tokens.
	Fusing places of different nets combines those nets into a single net of a more complex structure.
	
	This article uses a graphical representation of the net \pnHierarchical{} in which places are represented by single circles, pages by double circles, transitions by rectangles, directed arcs by arrows and tokens by black filled circles.
	Conditions \pnCondition{} associated with transitions are placed within square brackets.
	If a condition is always fulfilled (i.e.\ is True) then it may be omitted.
	Association of two places: $p_{\pnHierarchicalX{1},\alpha}^{\rm fusion}$ and $p_{\pnHierarchicalX{2},\beta}^{\rm fusion}$, belonging respectively to nets: \pnHierarchicalX{1} and \pnHierarchicalX{2} is represented by a single place $p_{(\pnHierarchicalX{1}, \pnHierarchicalX{2}),(\alpha,\beta)}^{\rm fusion}$.
	The definition of a safe HPN requires that in each of its places at the most one marker resides.

	\section{EMBODIED AGENT ARCHITECTURE}
	\label{sec:embodied-agent}
	
	\subsection{Structure}
\begin{figure}[!tbp]
		\centering
		\includegraphics[width=0.44\textwidth]{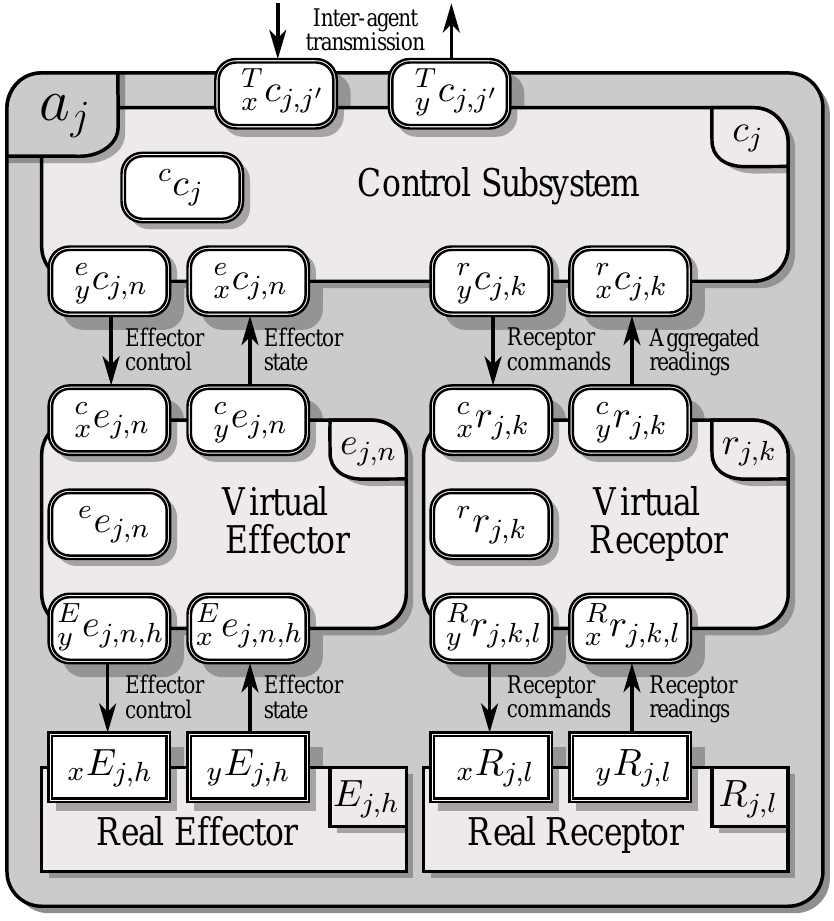}
		\caption{Internal structure of an embodied agent \agentj{j}}
		\label{fig:embodied_agent}
	\end{figure}

	The internal structure of an embodied agent \agentj{j} ($j$ -- name of the agent) is presented in Fig.~\ref{fig:embodied_agent}. An agent consists of the control subsystem \cbbbj{j} receiving the aggregated data about the environment from the virtual receptors \rbbbj{j,k} ($k$ -- name of a specific virtual receptor) and based on the received data and its internal imperative it formulates the control commands for its virtual effectors \ebbbj{j,n} ($n$ -- name of a specific virtual effector). Virtual effectors transform the commands received from \cbbbj{j} into a form  acceptable to the real effectors \Ebbbj{j,h} ($h$ -- name of real effector), whereas virtual receptors aggregate data from the real receptors \Rbbbj{j,l} ($l$ -- name of a real receptor) into a form acceptable to \cbbbj{j}. The agent's subsystems communicate with each other through buffers (as presented in Fig.~\ref{fig:embodied_agent}). A systematic method of naming buffers is used~\cite{Kornuta:13_irs,ZielinskiMMAR2014}. The letter in the center indicates the type of the subsystem $s$, where $\sbs \in \{\cbbbb, \ebbbb, \rbbbb,  \Ebbbb, \Rbbbb\}$. The right subscript determines the names of: the agent, the subsystem and if it is necessary, the buffer component. While, the left subscript determines, whether the buffer is an input buffer ($x$), output buffer ($y$) or internal memory (in this case this subscript is omitted). The right superscript determines the discrete time stamp, e.g.\ $i$, whereas the left superscript, determines the type of the subsystem, from which the data is obtained or to which the data is directed. For example: 1) \crixj{i}{j,v} is the content of the  input buffer receiving data from the virtual receptor output buffer at a discrete time $i$; this input buffer is a part of the control subsystem \cbbbj{j} of the agent \agentj{j}, 2) \ccibj{i}{j} is the internal memory of \cbbbj{j} at time $i$. It should be noted that each subsystem can run at a different sampling rate, thus $i$ for each of those subsystems is different (context distinguishes them).
	
	\subsection{Activities}
	\label{subsec:robotic-system-activity}
	
\begin{figure}
		\centering
		\includegraphics[width=0.5\textwidth]{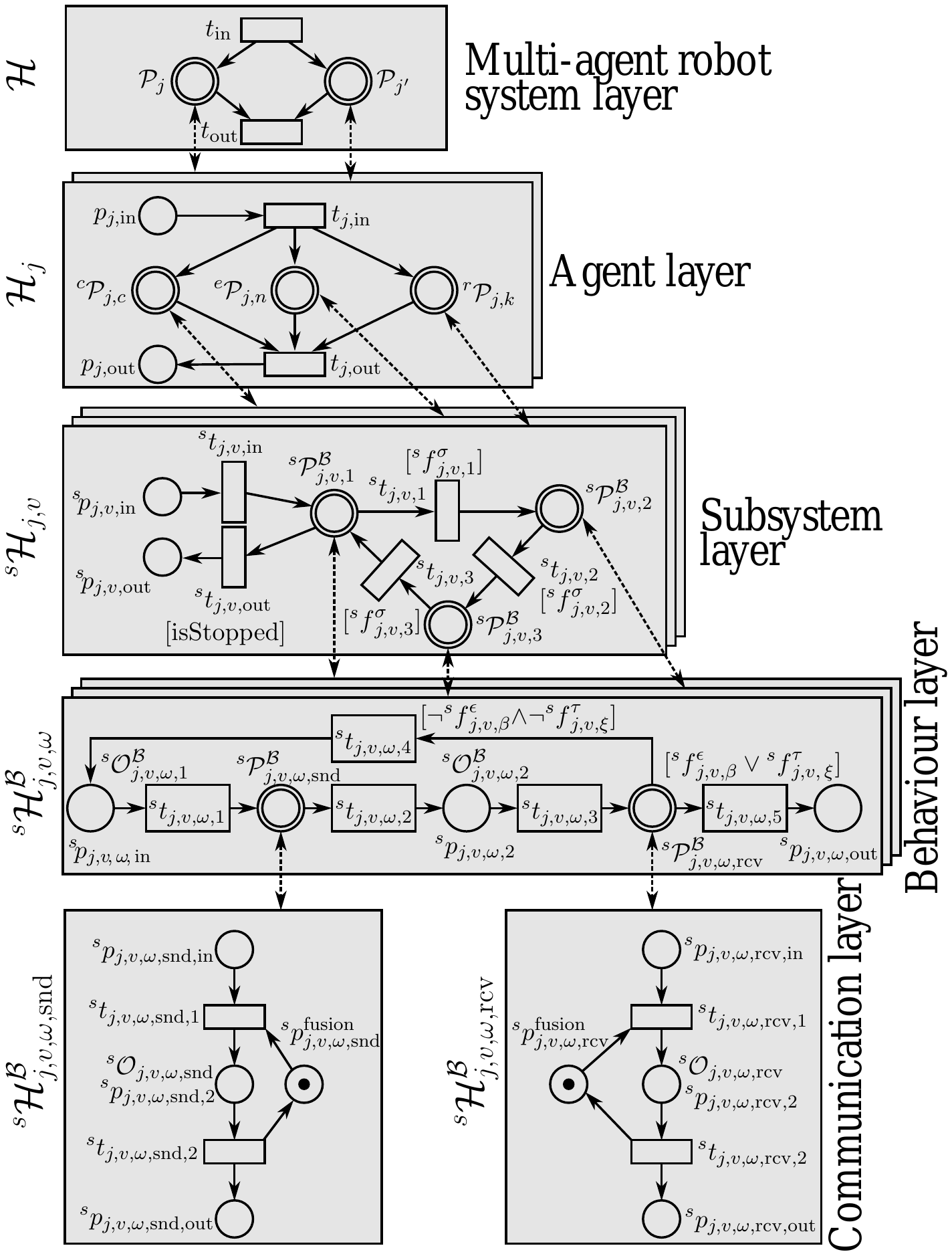}
		\caption{HPN \pnHierarchicalXXX{}{}{} defining the activities of a robot system}
		\label{fig:hierarchicalpetrinetconnections}
	\end{figure}

	A HPN \pnHierarchicalXXX{}{}{} determining the activities of a multi-agent robot system consists of five layers (Fig.~\ref{fig:hierarchicalpetrinetconnections}):
	\begin{enumerate}
		\item multi-agent robot system layer composed of a single net \pnHierarchicalXXX{}{}{} defining individual pages \pnPageXXX{j}{}{} for each agent \agentj{j} within the multi-agent robot system,
		\item agent layer composed of nets \pnHierarchicalXXX{j}{}{} responsible for the activation of pages  \pnPageXXX{j,v}{}{s}, which in turn determine the activities of individual subsystems \sbbbj{j,v} ($v$ is the subsystem designator), $\sbs \in \{\cbbbb, \ebbbb, \rbbbb\}$,
		\item subsystem layer defining nets \pnHierarchicalXXX{j,v}{}{s} represented by pages \pnPageXXX{j,v}{}{s}; each net \pnHierarchicalXXX{j,v}{}{s} contains pages \pnPageXXX{j,v\!,\,\omega}{\agentBehaviour}{s} defining particular behaviours \Bsbbj{j, v, \omega} of \sbbbj{j,v} ($\omega$ is the behaviour designator),
		\item behaviour layer defining behaviours \Bsbbj{j, v, \omega} represented by pages \pnPageXXX{j,v,\omega}{\agentBehaviour}{s},
		\item communication layer determining the communication models used by behaviours \Bsbbj{j, v, \omega}, i.e.\ defining pages \pnPageXXX{j, v, \omega, \rm snd}{\agentBehaviour}{s} (net describing behaviour of \sbbbj{j,v} when data is sent to other subsystems) and \pnPageXXX{j, v, \omega, \rm rcv}{\agentBehaviour}{s} (description of how \sbbbj{j,v} behaves when it receives data from other subsystems);  example is in Fig.~\ref{fig:communication-model-nonblocking-nonblocking}; It should be noted that the communication models used both for sending and receiving data pertain to the same subsystem \sbbbj{j,v}, and not to the system  \sbbbj{j,h} with which \sbbbj{j,v} communicates.
	\end{enumerate}
	
	Out of the five layers of \pnHierarchicalXXX{}{}{} three have user defined structure, the behaviour layer has a fixed structure, and the communication layer is composed of predefined blocks.
	The first two layers trigger parallel execution of nets representing the activities of agents and subsystems respectively. Subsystem \sbbbj{j,v} executes its behaviours \Bsbbj{j, v, \omega}.
	Any behaviour \Bsbbj{j, v, \omega}  calculates the transition function \sfbbbj{j,v,\omega} associated with it producing:
	\begin{equation}
	\left( \ssibj{i+1}{j,v}, \, \sbiyj{i+1}{j,v} \right) :=\ \sfbbbj{j,v,\omega} (\ssibj{i}{j,v},\, \sbixj{i}{j,v}),
	\label{eq:TransitionFunction}
	\end{equation}
	sends the produced output data (from the output buffer \sbbyj{j,v}) to the associated subsystems, updates the discrete time $i$, receives data from the associated subsystems (this data appears in the input buffer \sbbxj{j,v}) and finally checks the error condition \ecstbj{j,v,\beta} and terminal condition \tcstbj{j,v,\xi}, where $\beta$ and $\xi$ are the designators of the error and the terminal conditions respectively. The behaviour iterates until one of the above-mentioned conditions is fulfilled. Once a behaviour is terminated, the choice of the next one is based on a~predicate called an initial condition \icstbj{j,v,\alpha}($\alpha$ -- predicate designator)~\cite{zielinski:2016-automation}.
	Initial conditions of next behaviours are associated with the transitions pointed at by the arcs emerging from the place with which the just terminated behaviour is associated.
	

	
	The communication model must define whether communicating subsystems activities are blocked during the data transfer or not. Both the producer and the consumer may operate either in blocking or non-blocking mode. The blocking mode causes the producer to wait for the consumer to confirm that data has been received. In the non-blocking mode the producer does not wait for this confirmation and instantaneously resumes its other activities.  In the blocking mode the consumer waits until it receives the data, while in the non-blocking mode it resumes its other operations if data is unavailable.
	A non-blocking mode used by both the producer and consumer is called a fully asynchronous communication model.

	\section{CONSTRUCTION OF SYSTEM LAYERS}
	\label{sec:embodied-agent-style-by-petri-net}
	
	Construction of a HPN of consecutive system layers is performed as described below (Fig.~\ref{fig:hierarchicalpetrinetconnections}).
	
	\subsection{HPN \pnHierarchicalXXX{}{}{} of the multi-agent robot system layer}
	\begin{enumerate}
		\item Add to \pnHierarchicalXXX{}{}{} a page \pnPageXXX{j}{}{} for each agent \agentj{j} within the multi-agent robot system,
		\item Add to \pnHierarchicalXXX{}{}{} two transitions \pnTransitionXXX{\rm in}{}{} and \pnTransitionXXX{\rm out}{}{},
		\item Connect by directed arcs \pnTransitionXXX{\rm in}{}{} to each \pnPageXXX{j}{}{} and each \pnPageXXX{j}{}{} to \pnTransitionXXX{\rm out}{}{}.
	\end{enumerate}
	
	\subsection{HPN \pnHierarchicalXXX{j}{}{} of the agent layer}
	\begin{enumerate}
		\item Create a page \pnPageXXX{j,v}{}{s} for each subsystem \sbbbj{j,v} within \agentj{j}. For $\cbbbj{j}$ create page \pnPageXXX{j,c}{}{s},
		\item Create two transitions \pnTransitionXXX{j,\rm in}{}{} and \pnTransitionXXX{j,\rm out}{}{},
		\item Connect \pnTransitionXXX{j,\rm in}{}{} to each page by a directed arc,
		\item Connect each page to \pnTransitionXXX{j,\rm out}{}{} by a directed arc,
		\item Create a place \pnPlaceXXX{j,\rm in}{}{} and connect it to \pnTransitionXXX{j,\rm in}{}{} by a directed arc,
		\item Create a place \pnPlaceXXX{j,\rm out}{}{}; connect \pnTransitionXXX{j,\rm out}{}{} to \pnPlaceXXX{j,\rm out}{}{} by a directed arc.
	\end{enumerate}
	
	\subsection{HPN \pnHierarchicalXXX{j,v}{}{s} of the subsystem layer} 
	\begin{enumerate}
		\item Create a page \pnPageXXX{j,v,\omega}{\agentBehaviour}{s} for each behaviour \Bsbbj{j, v, \omega}
		exhibited by the subsystem \sbbbj{j,v},
		
		\item For each behaviour \Bsbbj{j, v, \omega} (i.e.\ \pnPageXXX{j,v,\omega}{\agentBehaviour}{s}) determine the initial conditions \icstbj{j,v,\alpha} that will be associated with transitions  \pnTransitionXXX{j,v,\alpha}{}{} leading to successor behaviours \Bsbbj{j, v, \delta} (i.e.\ \pnPageXXX{j,v,\delta}{\agentBehaviour}{s}). Connect \pnPageXXX{j,v,\omega}{\agentBehaviour}{s} to all successor pages in the following way: connect \pnPageXXX{j,v,\omega}{\agentBehaviour}{s} to \pnTransitionXXX{j,v,\alpha}{}{} and \pnTransitionXXX{j,v,\alpha}{}{} to \pnPageXXX{j,v,\delta}{\agentBehaviour}{s} by directed arcs for all successor pages \pnPageXXX{j,v,\delta}{\agentBehaviour}{s}.

		\item Add to \pnHierarchicalXXX{j,v}{}{s} two places: \pnPlaceXXX{j,v,\rm in}{}{s} -- input place and \pnPlaceXXX{j,v,\rm out}{}{s} -- output place, and two transitions \pnTransitionXXX{j,v,\rm in}{}{s} and \pnTransitionXXX{j,v,\rm out}{}{s}.  Place \pnPlaceXXX{j,v,\rm in}{}{s} is connected through \pnTransitionXXX{j,v,\rm in}{}{s} to a page that represents the initial behaviour of the subsystem. The page representing the terminal behaviour of the subsystem is connected by a directed arc to \pnTransitionXXX{j,v,\rm out}{}{s}, which in turn is connected by another arc to the place \pnPlaceXXX{j,v,\rm out}{}{s}. The places   \pnPlaceXXX{j,v,\rm in}{}{s} and \pnPlaceXXX{j,v,\rm out}{}{s} are the input and output places of \pnHierarchicalXXX{j,v}{}{s} (i.e.\ \pnPageXXX{j,v}{}{s}).

	\end{enumerate}

	\subsection{HPN \pnHierarchicalXXX{j,v,\omega}{\agentBehaviour}{s} of the behaviour layer}
	\label{subsec:behaviour-as-petri-net-page}
	
	The structures of all HPNs \pnHierarchicalXXX{j,v,\omega}{\agentBehaviour}{s} represented by \pnPageXXX{j,v,\omega}{\agentBehaviour}{s} follow a single pattern, thus they do not have to be created. Only the parameters of their activities have to be delivered, i.e.: transition function \sfbbbj{j,v,\omega}, error condition \ecstbj{j,v,\beta} and terminal condition \tcstbj{j,v,\xi}, as well as the  communication models utilised within pages \pnPageXXX{j,v,\omega,\rm snd}{\agentBehaviour}{s} and \pnPageXXX{j,v,\omega,\rm rcv}{\agentBehaviour}{s}. Behaviour \Bsbbj{j, v,\omega} is executed by \pnHierarchicalXXX{j,v,\omega}{\agentBehaviour}{s} in the following way.
	

	%

	\begin{enumerate}
		
		\item The execution of behaviour \Bsbbj{j, v, \omega} starts when a token appears in the place \pnInputPlaceXX{j,v,\omega,\rm in}{s}. This initiates the operation \pnOperationXXX{j,v,\omega,1}{\agentBehaviour}{s} which consists in calculation of  the transition function  \sfbbbj{j,v,\omega} by using (\ref{eq:TransitionFunction}). Transition \pnTransitionXXX{j,v,\omega,1}{}{s} will not fire until the calculation is complete.
		\item Firing the transition \pnTransitionXXX{j,v,\omega,1}{}{s} places a single token in the page \pnPageXXX{j,v,\omega,\rm snd}{\agentBehaviour}{s}, initiating its execution.
		Page \pnPageXXX{j,v,\omega,\rm snd}{\agentBehaviour}{s} represents a net \pnHierarchicalXXX{j, v,\omega,\rm snd}{\agentBehaviour}{s} determining the communication model used to send the contents of the output buffers \sbiyj{i}{j,v} to the connected subsystems.
		When the page \pnPageXXX{j,v,\omega,\rm snd}{\agentBehaviour}{s} finishes its execution, the transition \pnTransitionXXX{j,v,\omega,2}{}{s} is ready to fire,
		\item The operation \pnOperationXXX{j,v,\omega,2}{\agentBehaviour}{s} is associated with the place \pnPlaceXXX{j,v,\omega,2}{}{s} that receives the token. It increments the discrete time $i$ of the subsystem \sbbbj{j,v},
		\item Firing the transition \pnTransitionXXX{j,v,\omega,3}{}{s} activates the page \pnPageXXX{j,v,\omega,\rm rcv}{\agentBehaviour}{s}, which represents the net \pnHierarchicalXXX{j, v,\omega,\rm rcv}{\agentBehaviour}{s} determining in what mode is the data received from the associated subsystems,
		\item When the page \pnPageXXX{j,v,\omega,\rm rcv}{\agentBehaviour}{s} completes its activity, the error condition \agentErrorConditionS{j,v,\beta} and the terminal condition \agentTerminalConditionS{j,v,\xi} are checked.
		If none of them is fulfilled, the next iteration of behaviour \Bsbbj{j, v,\omega} starts. Otherwise a single token is placed in the output place \pnPlaceXXX{j,v,\omega,\rm out}{}{s\!}. This terminates the behaviour \Bsbbj{j, v,\omega} execution, and thus completes the activities of the page \pnPageXXX{j, v,\omega}{\agentBehaviour}{s}. The control returns then to the net \pnHierarchicalXXX{j,v}{}{s}.
		
	\end{enumerate}
	
	\subsection{Communication layer -- determination of communication model utilised within behaviour \agentBehaviourXXX{j,v,\omega}{}{s}}
	\label{subsec:communication-model}
	
	The nets composing the communication layer again do not have to be created. As the number of communication patterns is limited, only the selection of the required pattern has to be made. The communication layer is composed of two pages \pnPageXXX{j,v,\omega,\rm snd}{\agentBehaviour}{s} and \pnPageXXX{j,v,\omega,\rm rcv}{\agentBehaviour}{s}. The first one defines how \sbbbj{j,v} sends the data to other subsystems and the second one determines how it receives the data from other subsystems. Those two pages interact with similar pages of the subsystems that \sbbbj{j,v} communicates with. Those pages are created in the following way.
	\begin{enumerate}
		\item Select the communication model between subsystem \sbbbj{j,v} and other subsystems \sbbbj{j,h},
		\item Express this model in the form of a Petri net (selection of an appropriate net for the two subsystems \sbbbj{j,v} and \sbbbj{j,h}),
		\item Divide this Petri net into two Petri nets with fused places: a) presenting how the data is sent, b) presenting how the data is received,
	\end{enumerate}
	This procedure has to be repeated twice. Once to define how subsystem \sbbbj{j,v} sends the data and once to define how it receives data.
	
\begin{figure}[!tb]
		\fbox{\includegraphics[height=4.05cm]{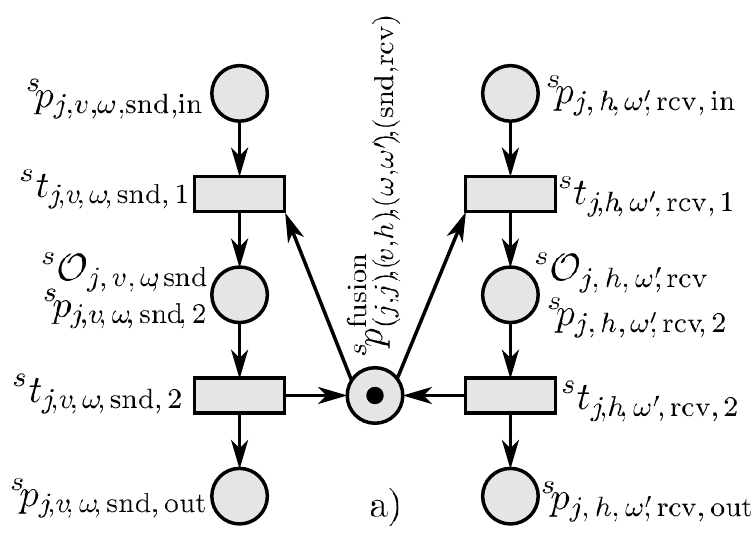}}
		\fbox{\includegraphics[height=4.05cm]{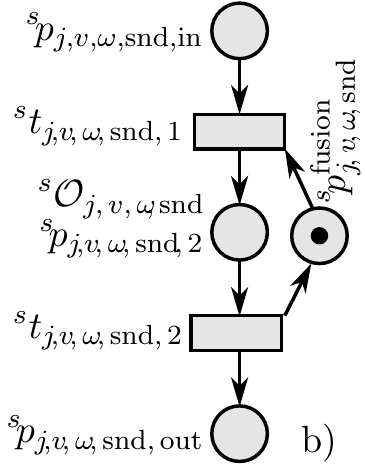}}
		\fbox{\includegraphics[height=4.05cm]{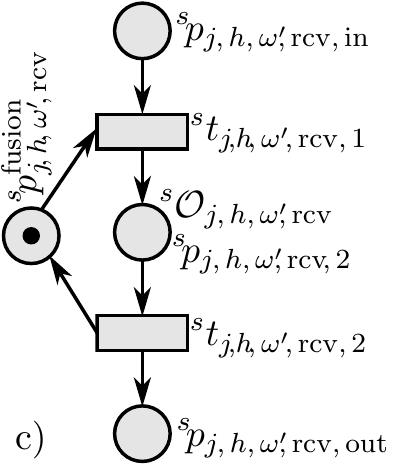}}
		\caption{{\bf (a)} Fully asynchronous communication model used to send data from subsystem \sbbbj{j,v} to subsystem  \sbbbj{j,h}.
			{\bf (b)} Page \pnPageXXX{j,v,\omega,{\rm snd}}{\agentBehaviour}{s} defines the communication model used by subsystem \sbbbj{j,v} executing behaviour \agentBehaviourXXX{j,v,\omega}{}{s}. {\bf (c)} Page \pnPageXXX{j,h,\omega',{\rm rcv}}{\agentBehaviour}{s} defines the communication model used by  subsystem \sbbbj{j,h} executing behaviour \agentBehaviourXXX{j,h,\omega'}{}{s}. Both pages are connected by the fusion place shown in (a).}
		\label{fig:communication-model-nonblocking-nonblocking}
	\end{figure}

	Out of several communication models the fully asynchronous one was selected to exemplify the design
	procedure. It functions in the following way. The two communicating subsystems are: \sbbbj{j,v} and \sbbbj{j,h}.  The former executes the behaviour \agentBehaviourXXX{j,v,\omega}{}{s} and sends data to the latter, which executes the behaviour \agentBehaviourXXX{j,h,\omega'}{}{s} (Fig.~\ref{fig:communication-model-nonblocking-nonblocking}a).
	Both subsystems, i.e.\ \sbbbj{j,v} and  \sbbbj{j,h}, in this mode of communication, compete for access to the communication channel, which contains shared memory. Only a single subsystem may access the shared memory at a time. When subsystem \sbbbj{j,v} gains access to the memory, it removes the token from the place \pnPlaceXXX{j,(v,h),(\omega,\omega'),(\rm snd,\rm rcv)}{\rm fusion}{s}, starts the activity associated with place \pnPlaceXXX{j,v,\omega,{\rm snd},2}{}{s} (inserts the data in the shared memory) and then inserts tokens in places \pnPlaceXXX{j,(v,h),(\omega,\omega'),(\rm snd, \rm rcv)}{\rm fusion}{s\!} and \pnPlaceXXX{j,v,\omega,{\rm snd},\rm out}{}{s}, thus terminating the communication episode. The system receiving the data acts similarly.
	The subscripts of the fusion place \pnPlaceXXX{j,(v,h),(\omega,\omega'),(\rm snd,\rm rcv)}{\rm fusion}{s\!} specify the direction of data transmission, i.e.\ from the subsystem \sbbbj{j,v} (executing behaviour \agentBehaviourXXX{j,v,\omega}{}{s}) to the subsystem  \sbbbj{j,h} (executing behaviour \agentBehaviourXXX{j,h,\omega'}{}{s}), emphasizing that this place belongs to both nets \pnHierarchicalXXX{j,v,\omega}{\agentBehaviour}{s} and \pnHierarchicalXXX{j,h,\omega'}{\agentBehaviour}{s}.
	The net presented in Fig.~\ref{fig:communication-model-nonblocking-nonblocking}a can then be decomposed into two nets that are respectively pages: \pnPageXXX{j,v,\omega,\rm snd}{\agentBehaviour}{s} (Fig.~\ref{fig:communication-model-nonblocking-nonblocking} b) and \pnPageXXX{j,h,\omega',\rm rcv}{\agentBehaviour}{s} (Fig.~\ref{fig:communication-model-nonblocking-nonblocking}c).
	Places \pnPlaceXXX{j,v,\omega,\rm snd}{\rm fusion}{s\!} and \pnPlaceXXX{j,h,\omega',\rm rcv}{\rm fusion}{s} are associated with each other creating a single place \pnPlaceXXX{j,(v,h),(\omega,\omega'),(\rm snd,\rm rcv)}{\rm fusion}{s\!}. Page \pnPageXXX{j,v,\omega,\rm rcv}{\agentBehaviour}{s} representing behaviour \agentBehaviourXXX{j,v,\omega}{}{s} is determined analogically.
	
	If the subsystem \sbbbj{j,v} sends or receives data to or from two or more subsystems, the pages of subsystem \sbbbj{j,v} have to be adequately connected.
	For instance, let subsystem \sbbbj{j,v} send data to  \sbbbj{j,h} and $s_{j,h'}$.
	Pages are defined for each pair of communicating subsystems.
	The pages \pnPageXXX{j,v,\omega,\rm snd, h}{\agentBehaviour}{s} and \pnPageXXX{j,v,\omega,\rm snd,h'}{\agentBehaviour}{s}, specifying the communication of  \sbbbj{j,v} with  \sbbbj{j,h} and $s_{j,h'}$, are connected with each other by association of output place of page \pnPageXXX{j,v,\omega,\rm snd, h}{\agentBehaviour}{s} with the input place of page \pnPageXXX{j,v,\omega,\rm snd, h'}{\agentBehaviour}{s}.
	This results in the creation of a single page \pnPageXXX{j,v,\omega,\rm snd}{\agentBehaviour}{s} describing sequential data transfer, first from \sbbbj{j,v} to  \sbbbj{j,h}, and then from \sbbbj{j,v} to $s_{j,h'}$. Alternatively, those pages can be connected in such a way that the data will be sent in parallel to the subsystems  \sbbbj{j,h} and $s_{j,h'}$.
	Other communication models mentioned in section~\ref{subsec:robotic-system-activity} require different Petri nets than the one presented in Fig.~\ref{fig:communication-model-nonblocking-nonblocking}a. They are not presented here due to lack of space.
	
	\section{EXAMPLE}
	\label{sec:experiment}
	The presented design methodology is currently tested by specifying robot systems and automatically generating their control code. The example presented here is purposefully kept  simple to explicate the methodology and not to obscure it by overly complex robot and its task. Thus the control system of a rudimentary simulated robot executing the follow the line task is presented.
	\subsection{Specification}
	
\begin{figure}[!tbp]
		\centering
		\begin{subfigure}[b]{0.20\textwidth}
			\includegraphics[height=2.6cm]{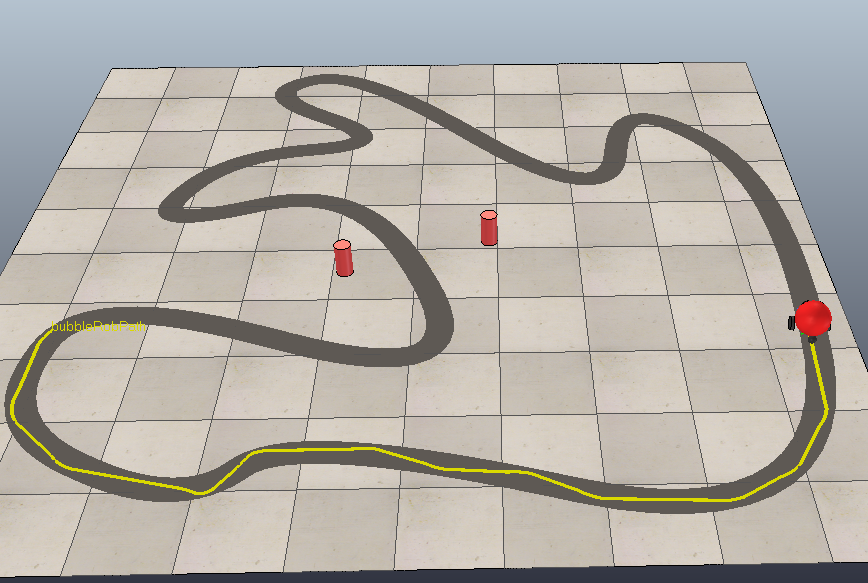}
			\caption{}
			\label{fig:simulation1}
		\end{subfigure}
		\hfill
		\begin{subfigure}[b]{0.23\textwidth}
			\includegraphics[height=2.6cm]{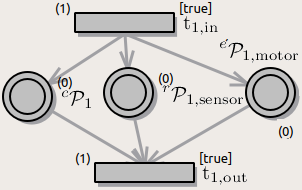}
			\caption{}
			\label{fig:agent-j-pn-tool}
		\end{subfigure}
		\caption{(a) Generated code simulation,
			(b) Agent layer net \pnHierarchicalXXX{1}{}{} designed using the tool developed by the authors}
		\label{fig:simulator1}
	\end{figure}

	The task of a two-wheeled robot is to follow a black curve painted on a plane contrasting surface (Fig.~\ref{fig:simulation1}).
	
	\paragraph{Robot system structure}
	The designed robot system is composed of a single embodied agent \agentj{1}. The agent \agentj{1} contains: control subsystem \cbbbj{1}, virtual receptor \rbbbj{1,\rm sensor}, virtual effector \ebbbj{1,\rm motor}, two real effectors \Ebbbj{1,\rm motor_1}, \Ebbbj{1,\rm motor_2}, being the two motors actuating two wheels, and three real receptors \Rbbbj{1,\rm sensor_1}, \Rbbbj{1,\rm sensor_2} and \Rbbbj{1,\rm sensor_3} detecting the intensity of the light reflected from the floor surface. The virtual receptor \rbbbj{1,\rm sensor} aggregates data from the real receptors and then forwards the result to \cbbbj{1}. The control subsystem, uses the data received from  \rbbbj{1,\rm sensor} and  sends  appropriate commands to \ebbbj{1,\rm motor}, which computes the inverse kinematics to control the real effectors.
	
\begin{figure}
		\centering
		\begin{subfigure}{.20\textwidth}
			\centering
			\includegraphics[height=6.4cm, clip]{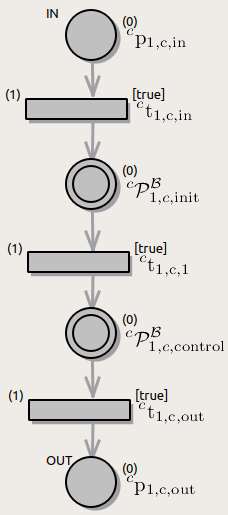}
			\caption{}
			\label{fig:cs-pn-tool}
		\end{subfigure}%
		\begin{subfigure}{.26\textwidth}
			\centering
			\includegraphics[height=6.4cm, clip]{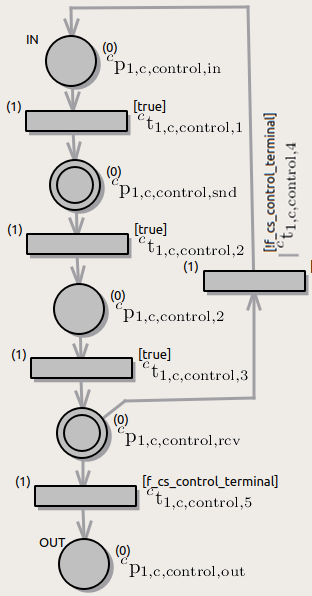}
			\caption{}
			\label{fig:c-control-pn-tool}
		\end{subfigure}%
		
		\caption{(a) \pnHierarchicalXXX{1}{}{c} net defining the subsystem layer of \cbbbj{1}, (b) \pnHierarchicalXXX{1,\rm control}{\agentBehaviour}{c} net defining the behaviour layer of \cbbbj{1} executing behaviour \agentBehaviourXXX{1,\rm control}{}{c}.}
		\label{fig:pn-tool-1}
	\end{figure}

	\paragraph{Robot system activities}
	The designed robot system consists only  of a single agent \agentj{1} thus HPN \pnHierarchicalXXX{}{}{} consists of a single page \pnPageXXX{1}{}{}.
	Petri net \pnHierarchicalXXX{1}{}{} determining the agent layer is presented in Fig.~\ref{fig:agent-j-pn-tool}, subsystem layer in Fig.~\ref{fig:cs-pn-tool}, behaviour layer in Fig.~\ref{fig:c-control-pn-tool} and communication layer in Fig.~\ref{fig:communication-model-nonblocking-nonblocking}a. For the sake of briefness this paper presents only the control subsystem \cbbbj{1} HPN  \pnHierarchicalXXX{1}{}{c}, which exhibits two behaviours: \agentBehaviourXXX{1,\rm init}{}{c}, which initiates the connection with the simulated robot, and \agentBehaviourXXX{1, \rm control}{}{c}, which calculates the transition function \tfebbj{1,\rm control} (Fig.~\ref{fig:transition-function-example}), i.e.\ calculates the robot linear ${\rm v_{\rm lin}}$ and angular  ${\rm v_{\rm ang}}$  velocities using the data received from \rbbbj{1,\rm sensor}, and sends them to \ebbbj{1,\rm motor}, where the wheel speeds are produced. The specification of \tfebbj{1,\rm control} (Fig.~\ref{fig:transition-function-example}) must be transformed manually into C++ code, which is subsequently inserted into the robot controller code by the development tool. \tfebbj{1,\rm control} is evaluated as an operation associated with the place \pnPlaceXXX{1,c,\rm control,\rm in}{}{c} in  Fig.~\ref{fig:c-control-pn-tool}.
	
	\begin{figure}
		\def\arraystretch{1.3}
		\[
		\begin{array}{l}
		\tfebbj{1,\rm control} \triangleq \\[2mm]
		\left\{
		\begin{array}{lcl}
		
		\left.\begin{array}{l}
		\ceiyj{i+1}{1,\rm motor}[{\rm v_{lin}}]={\rm v}\\	
		\ceiyj{i+1}{1,\rm motor}[{\rm v_{ang}}]={\rm 0}
		\end{array}\right\} &
		\mbox{for } & \mathcal{L} \land \mathcal{M} \land \mathcal{R}\\

		\left.\begin{array}{l}
		\ceiyj{i+1}{1,\rm motor}[{\rm v_{lin}}]={\rm v}\\	
		\ceiyj{i+1}{1,\rm motor}[{\rm v_{ang}}]={\rm -\omega/15}
		\end{array}\right\} &
		\mbox{for } & \mathcal{L} \land \mathcal{M} \land \lnot\mathcal{R}\\

		\left.\begin{array}{l}
		\ceiyj{i+1}{1,\rm motor}[{\rm v_{lin}}]={\rm v}\\	
		\ceiyj{i+1}{1,\rm motor}[{\rm v_{ang}}]={\rm -\omega}
		\end{array}\right\} &
		\mbox{for } & \mathcal{L} \land \lnot\mathcal{M} \land \lnot\mathcal{R}\\
		
		\left.\begin{array}{l}
		\ceiyj{i+1}{1,\rm motor}[{\rm v_{lin}}]={\rm v}\\	
		\ceiyj{i+1}{1,\rm motor}[{\rm v_{ang}}]={\rm \omega/15}
		\end{array}\right\} &
		\mbox{for } & \lnot\mathcal{L} \land \mathcal{M} \land \mathcal{R}\\
		
		\left.\begin{array}{l}
		\ceiyj{i+1}{1,\rm motor}[{\rm v_{lin}}]={\rm v}\\	
		\ceiyj{i+1}{1,\rm motor}[{\rm v_{ang}}]={\rm \omega}
		\end{array}\right\} &
		\mbox{for } & \lnot\mathcal{L} \land \lnot\mathcal{M} \land \mathcal{R}\\

		\left.\begin{array}{l}
		\ceiyj{i+1}{1,\rm motor}[{\rm v_{lin}}]={\rm -v}\\	
		\ceiyj{i+1}{1,\rm motor}[{\rm v_{ang}}]={\rm -\omega/2}
		\end{array}\right\} &
		& {\rm otherwise}\\

		\end{array}
		\right.
		\end{array}
		\label{TrFunc-ce}
		\]
		\def\arraystretch{1}
		\caption{Specification of \cbbbj{1} subsystem  partial transition function \tfebbj{1,\rm control}, where $\mathcal{L}\equiv\crixj{i}{\rm 1,\rm sensor}[{\rm left}]$, $\mathcal{M}\equiv\crixj{i}{\rm 1,\rm sensor}[{\rm middle}]$ and $\mathcal{R}\equiv\crixj{i}{\rm 1,\rm sensor}[{\rm right}]$ are binary signals produced by the left, middle and right sensor.}
		\label{fig:transition-function-example}
	\end{figure}

	\subsection{Petri net generation and execution}
	The generation of a robot controller code requires the creation of the \pnHierarchicalXXX{}{}{} HPN and its subnets. For that purpose a tool has been developed by the authors. The designed Petri net is automatically transformed into C++ code, forming the robot controller. The generated controller representing the HPN with an initial marking is then merged with the library executing the HPN. The resulting C++ code is compiled and the outcome is loaded into the control computer.
	The code invokes the scheduler, which searches for active transitions (enabled transition with fulfilled condition). One of such transitions is fired, i.e.\ a token is removed from each input place (place directly connected to the fired transition by a directed arc pointing at the transition) and inserts tokens into each output place (place directly connected to the transition by a directed arc pointing at the place). The operations associated with the output places are executed in separate threads. When the directed arc connects the firing transition with an output page, the input place of that page receives a new token and the associated operation of that input place starts its execution in a new thread. The scheduler repeats the above-mentioned steps either endlessly or until a behaviour commands it to terminate its activities.
	\subsection{Simulation}
	The robot following a line was simulated utilising the VREP simulator~\cite{vrep}.
	The simulator utilises a model of the robot (two-wheeled mobile robot from the VREP model library), exemplary environment (the floor with a black line on it) and the automatically generated robot controller, specified by the \pnHierarchicalXXX{}{}{} HPN. The generated controller is a C++ code which controls the execution of subsystems working in parallel: \cbbbj{1}, \rbbbj{1,\rm sensor}, \ebbbj{1,\rm motor}. Real subsystems \Ebbbj{1,\rm motor_1}, \Ebbbj{1,\rm motor_2},  \Rbbbj{1,\rm sensor_1}, \Rbbbj{1,\rm sensor_2} and \Rbbbj{1,\rm sensor_3} are a part of the VREP simulator.

	\section{RELATED WORKS}
	\label{sec:related-works}
	
	Non-hierarchical PNs have been used for robot control, e.g.~\cite{Zhou:1992}.
	Construction of HPNs by substituting transitions by PNs or places by PNs, and place or transition fusion is described in \cite{Huber91,Valk:2003}.
	Substitution of a PN for a place~\cite{Huber91} requires the determination of the input transitions and the output transitions in the parent PN.
	As the number of input places and output places in the substituted net is not limited, this unnecessarily complicates the structure of the net for our case.
	In~\cite{Vogler:1992,Zuberek:1996,Bluemke:1997,Luo:2015} a HPN is created by substituting transitions or places, however place or transition fussion is not considered, thus those nets do not meet the requirements imposed on nets modeling a robot system.
	In~\cite{Huber91} three types of place fusion are distinguished: instance fusion set, page fusion set and global fusion set. The fusion set is a set of places which can be fused into a single place. 
	The instance fusion set merges only places which are in the same page instance, the page fusion set merges places which occur within all instances of the same page, while the global fusion merges all places in the fusion set within the PN.  The proposed place fusion types are excessive for our purposes.
	
	As the above mentioned nets are too complex a version tailored to our purposes was developed for the description of the multi-agent robot system activities.
	Our HPN \pnHierarchical{} is created using only two constructs: substitution of a place by a net and global fusion of places.
	Moreover, unlike in \cite{Huber91,Vogler:1992,Bluemke:1997,Luo:2015}, where the substituting net could have any structure, a page in our approach can have only a single input place and a single output place. This causes that the definition of the page does not require reference to the transitions located in the parent net (because it is known with which transitions the place is connected).
	As a result, the page is much simpler to analyze.
	Moreover, in contrast to the above nets, the HPN we propose, associates operations with places and predicates with transitions.
	
	
	\section{CONCLUSIONS}
	\label{sec:conclusions}
	The paper presents a systematic methodology of designing multi-layer HPNs defining the activities of multi-agent robot systems.
	Our previous approach was based on HFSM. It treated the communication model used between subsystems as an implementation detail.
	The communicating subsystems acted independently of each other.
	Since embodied agents contain communicating subsystems and the communication model between those subsystems  was not revealed during system specification, it was difficult to verify the correctness of the developed robotic system. The interactions between subsystems using single HFSMs, assuming sequential execution of subsystem activities, was presented in~\cite{Zielinski-2017-JINT,Figat:2017_RoMoCo,Zielinski:Automation2018}.
	
	A natural inclusion of the communication model into the system specification is possible by using HPNs with conditions.
	The proposed approach, follows the primary principle of structured programming, stating that the programmer must keep at all times the produced code within his or her intellectual grasp~\cite{Dijkstra:1976}. It structures a robotic system into layers of PNs describing the activities of ever smaller modules.
	The presented approach enables automatic code generation of the robotic controller.
	The resulting specification is a single HPN describing the activities of the concurrently executed activities of subsystems, also describing their interactions.
	The proposed specification method can be used to describe the activities of any robotic system. For a multi-robot system the developer defines separate HPNs for each robot (being itself a multi-agent system). Those HPNs communicate with each other using communication models implemented as interprocess communication.
	
	The purpose of the example presented in this paper has been the exemplification of the proposed specification and implementation method by focusing the reader's attention only on the methodology, treating the specified robotic system as of secondary importance, thus its rudimentary character.
	More complicated systems require the development of more complex HPNs.
	Any addition of a place to a net requires 6 extra lines of code, a new transition generates 6 additional lines, an extra directed arc only a single line, while a page 48 lines.
	Thus it can be shown that the size of the generated code grows linearly with  the size of the designed Petri net, hence the presented  approach based on HPNs is scalable.


	

	

\end{document}